\documentclass[]{article}

\pdfoutput=1

\usepackage[affil-it]{authblk}
\usepackage{natbib}
\usepackage{amssymb}
\usepackage{amsmath}
\usepackage{amsfonts}
\usepackage{calligra}
\usepackage{calrsfs}
\usepackage{hyperref}
\usepackage{url}
\usepackage[toc, page]{appendix}
\usepackage{booktabs}
\usepackage{float}
\usepackage{graphicx}
\usepackage{caption}
\usepackage{subfig}
\usepackage{geometry}
\usepackage[mathcal]{euscript}

\newcommand{\R}{\mathbb{R}}
\newcommand{\setsize}[1]{\left\vert{#1}\right\vert}
\newcommand{\OneHot}{\textbf{o}_w}
\newcommand{\NoteBOW}{\textbf{n}_j^{(i)}}
\newcommand{\Note}{\boldsymbol\eta_j^{(i)}}
\newcommand{\PatientData}{\boldsymbol\Pi^{(i)}}
\newcommand{\PatientDataLong}{\{ \Note \}_{j = 1}^{T^{(i)}} }

\newcommand{\PatientRepr}{\textbf{p}^{(i)}}
\newcommand{\XBOW}{\textbf{X}_{BoW}}
\newcommand{\PatientBOW}{\PatientRepr_{BoW}}
\newcommand{\PatientLSA}{\PatientRepr_{LSA}}
\newcommand{\NoteLDA}{\textbf{t}_j^{(i)}}
\newcommand{\PatientLDA}{\PatientRepr_{LDA}}
\newcommand{\PatientEA}{\PatientRepr_{EA}}
\newcommand{\PatientEAmean}{\PatientRepr_{EA_{mean}}}
\newcommand{\PatientEAmax}{\PatientRepr_{EA_{max}}}
\newcommand{\PatientRNN}{\PatientRepr_{RNN}}
\newcommand{\Vocab}{\mathcal{V}}
\newcommand{\TrainData}{\mathcal{P}_T}
\newcommand{\AllData}{\mathcal{P}}
\newcommand{\Embd}{\textbf{e}_w}
\newcommand{\SumOverNotes}{ \sum_{j = 1}^{T^{(i)}} }

\title{Efficient Representations of Clinical Text}

\author[1]{Sebastien Dubois}
\author[1]{Nathanael Romano}
\author[2]{David Kale}
\author[1]{Nigam H Shah}
\author[1,*]{Kenneth Jung}

\affil[1]{Stanford Center for Biomedical Informatics Research}
\affil[2]{USC Information Sciences Institute, University of Southern California}
\affil[*]{Corresponding author: kjung@stanford.edu}

\begin{document}
\maketitle

\begin{abstract}
Clinical notes are a rich source of information about patient state.  However, using them to predict clinical events with machine learning models is challenging.  They are very high dimensional, sparse and have complex structure.  Furthermore, training data is often scarce because it is expensive to obtain reliable labels for many clinical events.  These difficulties have traditionally been addressed by manual feature engineering encoding task specific domain knowledge.  We explored the use of neural networks and transfer learning to learn representations of clinical notes that are useful for predicting future clinical events of interest, such as all causes mortality, inpatient admissions, and emergency room visits.   Our data comprised 2.7 million notes and 115 thousand patients at Stanford Hospital.  We used the learned representations, along with commonly used bag of words and topic model representations, as features for predictive models of clinical events.  We evaluated the effectiveness of these representations with respect to the performance of the models trained on small datasets.  Models using the neural network derived representations performed significantly better than models using the baseline representations with small ($N < 1000$) training datasets.  The learned representations offer significant performance gains over commonly used baseline representations for a range of predictive modeling tasks and cohort sizes, offering an effective alternative to task specific feature engineering when plentiful labeled training data is not available. 
\end{abstract}

\section{Introduction}
Our goal is to devise a representation of clinical notes that allows us to make accurate predictions of a wide variety of future events, such as all causes mortality, inpatient admissions, and emergency room visits.  Clinical notes record important details about patients that cannot be found elsewhere in electronic health records (EHRs), but they present challenges to health care researchers, including very high dimensionality, sparsity, and complex linguistic and temporal structure.  Furthermore, reliable labels for many clinical outcomes are scarce relative to the complexity of the data because the gold standard for such labels remains manual review of patient records by clinicians.  We are thus specifically interested in representations of clinical notes that are \emph{efficient}, i.e., they admit good performance even labeled data is scarce.  Traditionally, these challenges are met by extensive feature engineering, guided by task specific domain knowledge.  We want to reduce the need for such feature engineering by automatically learning how to summarize a sequence of clinical notes about a given patient in a data driven manner.  The learned patient-level representation of the clinical notes should directly mitigate or admit simple strategies for overcoming the above challenges.  Specifically, it should support accurate models learned from limited training data, and  be useful across a wide range of clinical prediction tasks and research questions with minimal or no task-specific customization.  

We take a transfer learning approach to learning this representation.  Transfer learning has been used to achieve state of the art results in computer vision and Natural Language Processing (NLP) on tasks with limited labeled training data.  Such tasks can be challenging for very flexible models such as deep neural nets.  Transfer learning addresses this problem by first training a large, complex model on a \emph{source task} with a large dataset that has abundant labels.  The resulting model, in the course of learning how to perform the source task, learns a transformation of the input data into a useful form, i.e., \emph{a representation}, that allows it to solve the source task well.  We then use this model on data for a \emph{target task} with more limited data.  We hope that the learned representation is more effective for solving the target task than the original, raw input.  

We apply this paradigm to learning efficient patient-level representations of clinical notes.  \emph{Efficient} in this study means specifically that the representations allow us to perform well on the target task even when we have very limited training data.  \emph{Patient-level} means that our method must summarize the variable length sequence of clinical notes for a given patient into a fixed length vector suitable for use by simple models such as logistic regression.  We take two approaches, both based on commonly used techniques in Natural Language Processing (NLP).  The first approach, \emph{embed-and-aggregate}, uses GloVe \citep{pennington2014glove} to learn vector space representations, or embeddings, for biomedical concepts mentioned in clinical text.  We combine the embeddings for the concepts in each note into note representations using simple arithmetic operations.  These note representations are then similarly combined into patient representations.  In the second approach, we use recurrent neural nets (RNNs) \citep{elman1990finding} to learn representations of concepts, notes, and patients simultaneously.  In this approach, we must provide some form of supervision for the source task; we use contemporaneous data from another data modality, i.e., discrete diagnoses codes, as labels for each patient.  The input for both methods is a large corpus of clinical notes from Stanford Hospital.  

The resulting representations are evaluated by using them as features for predictive models for complex clinical outcomes -- all causes mortality, inpatient admissions, and emergency room (ER) visits.  For each representation, we estimate \emph{learning curves}, i.e., the relationship between model performance and training set size, because we are particularly interested in the utility of these features in situations where labeled training data is scarce.  We find that the learned representations performed significantly better than common baseline representations based on bag of words (BOW) and topic models for small labeled training sets ($N < 1000$ patients). 

\section{Background}
Let $\AllData$ be the set of all patients, indexed by $i$.  For each patient, we have a sequence of notes, $\PatientData = \PatientDataLong$, with each note $\Note$ indexed by $j$ in order from oldest to most recent, and with $T^{(i)}$ the number of notes for patient $i$.  Ultimately, our goal is to learn a function that estimates the probability of some clinical event of interest, such as all causes mortality, given the sequence of notes for a patient: 

\[ F\left( \PatientData \right) \approx Pr( \text{event of interest} \mid \PatientData) \]

\begin{figure}[h]
\centering \includegraphics[width=0.6\textwidth]{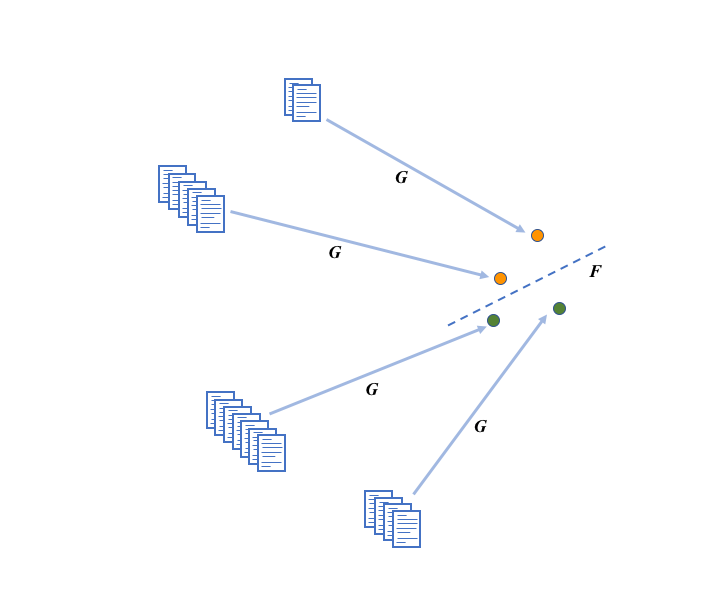}
\caption{Transfer learning decomposes our learning problem into two steps.  First, we we use unlabeled data to learn a function $G$ that maps variable length sequences of clinical notes into a fixed sized vector space.  Then we learn the function $F$ for our target task in that vector space using labeled data.}
\label{fig:problem_decomposition}
\end{figure}

However, $\PatientData$ is quite complex.  Each patient has a variable number of notes, so the size of the input to $F$ is not uniform.  Furthermore, each note, $\Note$, is itself a variable length sequence of words from a large vocabulary $\Vocab$.  And finally, in the clinical domain we often have very limited labeled datasets for supervised learning, i.e., we have  
$\left\{ ( \PatientData, y^{(i)} ) \right\}_{i \in \TrainData}$, with $\setsize{\TrainData} \ll \setsize{\AllData}$.  
In such situations, it is hard to learn $F$ directly as a mapping from the raw notes to probabilities.  Instead we can devise a transformation of $\PatientData$ into an easier-to-use form, e.g., a fixed length vector of real numbers suitable for use as features for off the shelf models such as logistic regression or gradient boosted trees, i.e., 
\[ G: \PatientData \rightarrow \R^d \]
We refer to the transformed data as the patient representation, $\PatientRepr = G\left( \PatientData \right), \PatientRepr \in \R^d$, and we generally aim to have $d \ll \setsize{\Vocab}$ to achieve dimensionality reduction in order to make the learning problem more tractable \footnote{Note that the transformation can be explicitly specified, i.e., $G$ can be the result of manual feature engineering.  However, in this work we focus on transformations that are learned from the data instead.}.  We can now try to learn: 
\[ F(\PatientRepr) = F\left( G\left( \PatientData \right) \right) \approx Pr\left( \text{event of interest} \mid \PatientData\right) \]
We have decomposed the problem into two steps: estimating $G$ using plentiful data, and then estimating $F$ using our limited, labeled data (\autoref{fig:problem_decomposition}).  The key insight in transfer learning is that even when $\TrainData$ is small, we can sometimes learn an excellent $G$ using the much larger dataset $\AllData$ \footnote{Note that in order to map $\PatientData$ into $\R^d$ with $d \ll \setsize{\Vocab}$, we need to aggregate information across both within and between notes.}.

\noindent In this section, we review the most commonly used methods for devising the transformation $G$, and briefly review their shortcomings.  We then introduce our proposed methods.  

\subsection{Bag of Words Representations and their shortcomings}
Our starting point is the \emph{one-hot} representation of each word $w$ in our vocabulary $\Vocab$.  Each $w$ is associated with a unique index $i_w \in \{ 1, 2, \dots , \setsize{\Vocab} \}$, and we represent each $w \in \Vocab$ with a vector $\OneHot \in \{0, 1\}^{\setsize{\Vocab}}$, with a $1$ at position $i_w$ and $0$'s elsewhere.  The \emph{bag of words} representation of a note is simply the sum of the one-hot vectors of the words appearing in that note -- for the $j$-th note for patient $i$: 
\[ \NoteBOW = \sum_{v \in \text{ note j }} \OneHot \]
We now define our baseline patient representation to be the patient-level bag of words: 
\[ \PatientBOW = \SumOverNotes \NoteBOW \]

Bag of words representations are simple and transparent but have shortcomings inherited from one-hot representation of words.  They retain the high dimensionality and sparseness of one-hot word vectors.  More critically, they violate the intuition that semantically similar notes should have similar representations.  Since we are representing words and notes as vectors, we would like similar words and notes to be close together in that space. For instance, the one hot representations for \emph{non-small cell lung cancer} and \emph{lung adenocarcinoma} are as unrelated to each other as they are to \emph{hypertension}.  And since our baseline note level representation is the sum of the one-hot vectors for the concepts appearing in the note: $\NoteBOW = \sum_{v \text{in note}} \OneHot$, it inherits these problems -- \emph{Tarceva} is the same drug as \emph{erlotinib}, but our bag of words representations of \emph{Received Tarceva for non-small cell lung cancer} is as distant from \emph{lung adenocarcinoma was treated with erlotinib} as it is from \emph{the quick brown fox}.  Our baseline patient representation inherits these problems, albeit possibly to a lesser extent since they aggregate information across many notes.  These disadvantages are exacerbated when we attempt supervised learning with small datasets because some variants of semantically similar words may not appear in the training data at all.    

\subsection{Improving on the Bag of Words Representation}
Bag of words representations are simple and can be quite effective when we have enough training data.  However, $\PatientBOW$ is still very high dimensional and sparse.  Many researchers thus reduce the dimensionality of $\PatientBOW$ to arrive at a final $\PatientRepr$.  We describe here two simple, widely used methods.  Latent Semantic Analysis (LSA), a classic method from information retrieval, uses a low rank approximation of the matrix $\XBOW \in \R^{ \setsize{\Vocab} \times \setsize{\mathcal{P}} }$, in which the $i$-th row is $\PatientBOW$ as defined above.  Specifically, it calculates a truncated singular value decomposition of $\XBOW$: 
\[ \XBOW \approx \textbf{U} \Sigma \textbf{V}^T  \]
$\textbf{V} \in \R^{\setsize{\Vocab} \times K}$ is an orthogonal, rank $K$ matrix that can use used obtain a $K$ dimensional patient representation, $\PatientLSA = \textbf{V}^T \PatientBOW \in \R^K$.  In practice, practitioners often use some transformation of $\XBOW$, such as some variation of the \emph{Term Frequency / Inverse Document Frequency} transformation, that weights each word in an unsupervised manner.  

Topic models such as Latent Dirichlet Allocation \citep{blei2003latent} are often presented as a probabilistic version of LSA that models each note as a categorical distribution over $K$ latent factors, interpreted as \emph{topics}.  Thus, each note is transformed: 
\[ \Note \rightarrow \NoteLDA \in [\theta_{jk}^{(i)}]_{k=1}^K \text{ where } \sum_{k = 1}^K \theta_{jk}^{(i)} = 1 \text{ and } \theta_{jk}^{(i)} \ge 0 \text{ }  \forall \text{ } k \]
In order to arrive at a patient level representation, we aggregate the topic distribution for each of a patient's notes.  In \emph{DeepPatient} \citep{miotto2016deep}, this was done by simply averaging the topics over each patient's notes: 
\[ \PatientLDA = \frac{1}{T^{(i)}}\sum_{j=1}^{T^{(i)}} \NoteLDA \]
Both of these approaches reduce the dimensionality of our patient representations to $K \ll \setsize{\Vocab}$, and are \emph{unsupervised} -- we do not need patient labels so we can use $\AllData$ to learn these transformations.  

\subsection{Word embeddings for clinical concepts}
Many of the weaknesses of bag of words representations are inherited from the implicit one-hot encoding of words.  There are two main approaches to overcoming these shortcomings.  Some researchers use explicit domain knowledge, e.g., hierarchies of concepts encoded in biomedical ontologies.  Alternatively, using techniques developed in NLP, we can learn better concept representations in a data driven manner.  Specifically, for each $w \in \Vocab$ we learn \emph{vector space embeddings} in which each $w$ is represented by a fixed length vector of real numbers, $\Embd \in \R^d$, with $d \ll \setsize{\Vocab}$.  These embeddings are learned from large, unlabeled corpora such that semantically similar words have similar $\Embd$.  The precise meaning of \emph{similar} varies for different algorithms for learning these embeddings \citep{pennington2014glove,mikolov2013w2v}, but the essential idea is to learn embeddings that embody the distributional hypothesis (Harris, 1954) -- similar words appear in similar contexts.  In NLP, it has been found that simply averaging the embeddings for words appearing in a document is an excellent document level representation for many tasks \citep{Pagliardini2017, Wieting2015, Shen2018, Arora2017iclr}.  We thus hypothesize that a similar approach, applied over concepts appearing in a note, and then over the series of notes for a patient, may be an effective patient representation.  For instance, we may take: 
\[ \PatientEAmean = \frac{1}{T^{(i)}} \sum_{j = 1}^{T^{(i)}} \frac{1}{\text{num words in note}} \sum_{ w \text{ in note } j} \Embd \]
i.e., we average the embeddings in each note and then across the notes for a given patient.  Note that we could have chosen a different aggregation function, for instance, an element-wise max: 
\[ \PatientEAmax =  \underset{j = 1 \dots T^{(i)}}{max} \underset{ w \text{ in note } j}{max} \Embd \]
In our experiments we focus on the \emph{min}, \emph{mean} and \emph{max} functions applied element-wise to word and note level representations.  Details for this approach, which we refer to as \emph{embed-and-aggregate}, in provided Materials and Methods.  

\subsection{Transfer learning with Recurrent Neural Nets}
Recurrent Neural Nets (RNNs) are widely used to model variable length sequences, and have been widely applied to both general EHR and ICU time series \citep{choi2016learning,lipton2015learning,rajkomar2018}.  However, these prior works directly train models for the target tasks of interest, and we are primarily interested in target tasks for which labeled training data is very limited.  We thus take a transfer learning approach in which we train an RNN on a \emph{source task} devised from $\AllData$.  In order to solve the source task, the RNN learns to transform variable length input sequences, each element of which is a bag of words corresponding to the notes about a given patient on a particular day, into a fixed length vector, $\PatientRNN \in \R^d$, suitable for the source task. Our transformation function $G$ is embodied by the RNN.  We then use $G$ on patients in $\TrainData$, which has labels for the \emph{target task} we are actually interested in, to learn $F$.  We provide details for this approach in Materials and Methods.  

\section{Related Work}

There has been a great deal of research showing the utility of clinical text for a variety of applications, including pharmacovigilance ~\citep{lependu2013pharmacovigilance}, coding~\citep{perotte2014diagnosis}, predictive modeling ~\citep{ghassemi2014unfolding,yu2015jamia}, and phenotyping~\citep{ford2016text}. Bag of Words is often the standard representation of choice, though often in combination with substantial preprocessing. Given the difficulties in working with clinical text it is desirable to minimize the amount of processing needed of the source textual document.

For example, \citep{jung2015functional} found that across a number of clinical data mining questions, using counts of present, positive mentions of medical terms yielded comparable performance to a more computationally intensive NLP pipelines that performed syntactic parsing and part-of-speech tagging. Besides bag of words, topic models such as latent Dirichlet allocation \citep{blei2003latent} have been a popular choice for learning low-dimensional representations of clinical text. Halpern et al \citep{halpern2012comparison} found that topic models underperformed BOW when used as features in a prediction model. Chen et al \citep{chen2015bridges} showed a high level of agreement between topic models trained on corpora from different institutions, indicating that they may provide a generic representation that can cross boundaries. Our results using embeddings trained on abstracts of biomedical journal articles \citep{de2014medical} suggest the same may be true of embeddings.

In other related work, Miotto et al, \citep{miotto2016deep} use a topic model to pre-process bag of words derived from clinical notes before combining them with counts over structured clinical data. This combined representation is then fed into an autoencoder, followed by a random forest classifier to predict future diagnoses. Such a combined pipeline makes it difficult to evaluate the individual contributions of bag of words or topic models.

Inspired by recent successes of word embedding models like word2vec~\citep{mikolov2013distributed} and GloVe~\citep{pennington2014glove}, a number of health care researchers have sought to learn embeddings of medical words or concepts~\citep{minarro2013exploring}. \citep{de2014medical} trained the embeddings derived from journal abstracts, while \citep{choi2016learning} focused on designing quantitative evaluations of how well embeddings capture known medical concepts. None of this prior work, evaluated the utility of their embeddings as features in predictive models.

Liu et al, \citep{liu2016transferring} described an innovative way to leverage embeddings to perform transfer learning in low data regimes. They provide a useful initial bias by rescaling each feature proportional to its similarity with the prediction target based on text descriptions of each.

There is a growing interest in directly modeling sequential clinical data without ad hoc feature engineering. Lipton et al, \citep{lipton2015learning} and Choi et al, \citep{choi2015doctor} apply RNNs to time series of physiology and codes, respectively, while \citep{razavian2016multi} applies a convolutional network to longitudinal lab results.
Several recent works have modeled sequences of clinical notes as time series of topics using Gaussian processes~\citep{ghassemi2014unfolding,ghassemi2015multivariate}, or dynamic Bayesian networks 
~\cite{caballero2015dynamically}.

Our work stands out in two ways: first, we focus on evaluating the utility of alternative representations of textual content for building predictive models specifically when training sets are small. Second, we find that, consistent with recent work in NLP, a conceptually simple approach, embed-and-aggregate, is competitive with more elaborate representation learning methods when building predictive models.

\section{Methods and Materials}

\subsection{Data}
Our data set comprises EHR data collected over a five year period (2009 through 2014) at Stanford Hospital.  These records were extracted into the Stanford Translational Research Integrated Database Environment (STRIDE) \citep{lowe2009stride}, a clinical data warehouse, where they were de-identified before being made available for research use.  The Stanford School of Medicine IRB determined that the use of this data was non-human subjects research.  Each patient's data comprised a timestamped series of encounters with care providers over the five year period.  Encounters generated both structured and unstructured data.  The structured data typically consists of discrete codes from controlled vocabularies for diagnoses, medications, and procedures, along with quantitative measurements such as vitals and laboratory measurements.  The unstructured data consists of the free text clinical notes written by care providers documenting each encounter and other aspects of patient care.  Our study focuses on the use of the clinical notes to predict clinical events.  

\subsubsection{Text Processing}
The text of the clinical notes was processed using the methods described in \citep{LePendu2013CPT}.  Briefly, we find occurrences of terms in an expansive vocabulary of biomedical terms compiled from a collection of controlled terminologies and biomedical ontologies.  Each term mention with is also tagged with negation (e.g., \emph{atrial fibrillation was ruled out}) and family history (e.g., \emph{The patient's father had hypertension}).  Each note is thus reduced to a bag of words representation, and any terms that are not recognized as biomedical concepts are discarded, along with the sequence of words appearing in a note.  Each term is mapped to unique biomedical concepts \citep{LePendu2013CPT} using a mapping based on the the Unified Medical Language System (UMLS) Metathesaurus, which provides a mapping of strings to Concept Unique Identifiers (CUIs) \citep{Bodenreider2004}.  We retained negated concepts as distinct elements in our vocabulary because negative findings may be informative in clinical medicine.  Finally, we removed concepts that appeared in fewer than 50 or more than 10 million notes.  This resulted in a final vocabulary of 107,388 concepts.  For the remainder of this paper, we refer to concepts as "words" for convenience.  We emphasize that as a result of this text processing, all of our representations are on an equal footing in that they are all starting from a bag of words occurring in each note.  We discuss the implications of this choice, along with other choices in how the text is handled, in the discussion section.  

\begin{figure}[h]
\centering \includegraphics[width=0.9\textwidth]{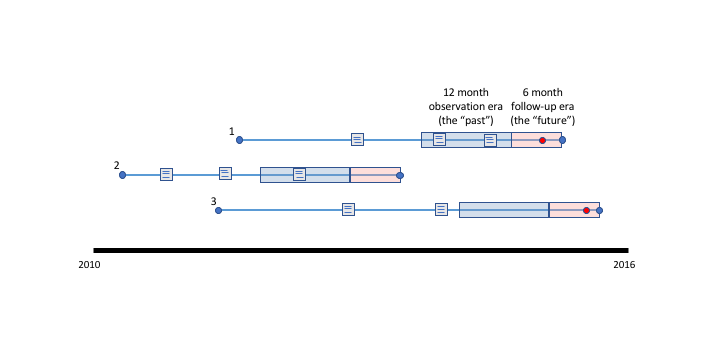}
\caption{Schematic for how we constructed a prediction problem from retrospective data.  The figure shows three patients, numbered 1-3.  Each patient's data consists of longitudinal observations (blue lines) comprising both structured (e.g., diagnosis codes) and unstructured data (free text of clinical notes, shown as gray rectangles on each timeline).  For each patient, we define the prediction time as six months prior to the end of their records in the EHR, which may correspond to the last observed clinical encounter or mortality. Note that patient timelines are not synchronized; each may have a different start and end, and thus prediction time.  We define the twelve months prior to each patient's prediction time the observation era (shown in blue on each patient timeline); this is "the past" from which we draw clinical notes to use to predict clinical events (red dots) in the follow-up era (shown in pink on each patient timeline), which is "the future" with respect to the prediction time. }
\label{fig:prediction_times}
\end{figure}

\subsubsection{Prediction times and data splits}
For simplicity, we set the prediction time for each patient as six months prior to the end of their record, where the end of a patient's record is determined by a recorded clinical encounter or mortality event as ascertained from the Social Security Death Index, hospital records, or state death register records.  We refer to the 12 months prior to the prediction time as the \emph{observation era}, and the 6 months after the prediction time as the \emph{follow-up era}; we use clinical notes from each patient's observation era to make predictions about whether or not clinical events occur in the patient's follow-up era (\autoref{fig:prediction_times}) Note that this choice of prediction time likely induces some bias in the learning problem as in actual usage we of course do not make predictions relative to some uncertain future event (i.e., a clinical encounter of mortality). However, the focus of our study is not on absolute performance but the relative performance of different strategies for representing clinical notes.

After filtering out patients who do not have any clinical notes with recognized biomedical concepts in their observation era, we were left with 115,232 patients, with 2,735,647 notes.  These patients were split randomly into training, validation and test sets with 69,417, 11,290, and 34,525 patients, respectively.  The validation set was used in preliminary experiments to tune the representation learning methods; all final evaluations were conducted on the hold out test set.  

\subsubsection{Target tasks}
We focus on clinical events that are complex, noisy functions of the input data: six month mortality, emergency department visits, and inpatient admissions.  These labels are ascertained from patient demographic and encounter data from the EHR. Note that these targets are low prevalence, so these problems exhibit significant class imbalance -- the prevelances of the outcomes are 2.8\%, 8.5\%, and 5.1\% respectively.  

\begin{figure}[h]
\centering \includegraphics[width=0.9\textwidth]{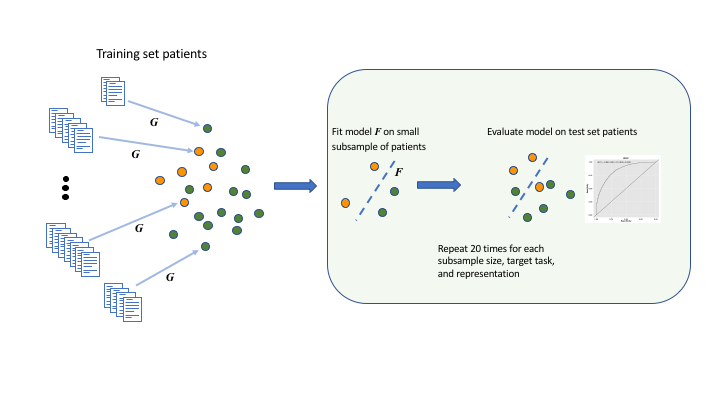}
\caption{Experimental Setup.  For each $G$ that we devise to transform variable length sequences of clinical notes into a patient level representation, we perform the following experiment 20 times: for each target task and subsample size (n = 125, 250, 500, 1000, 2000, and 4000), we subsample the \emph{source task training set} patients to get a \emph{target task training set} for the target task, balanced to 20\% positive case prevalence.  Then we fit a predictive model on the subsample and evaluate it on the test set (with the natural positive case prevalence).  }
\label{fig:expt_setup}
\end{figure}

\subsection{Experimental setup}
We are interested in how our representations do when we have only a small set of labeled examples for the target task, along with a much larger dataset of unlabeled examples. Our experimental setup is thus as follows \autoref{fig:expt_setup}.  First, we define the \emph{source task training set} as the data from the observation era of the 69,417 training set patients, i.e., $\AllData$ as defined above.  We use this data to learn patient level representations (i.e., the transformation function $G$).  Note that no information from the follow-up eras of these patients is used to learn $G$.  Next we define the \emph{target task training set}, i.e., $\TrainData$ as defined above, to be a random subset of the source task training set, plus labels derived from clinical events in those patients' follow-up eras. We use this data, transformed by $G$ into patient level representations, to learn $F$.  We vary the size of $\TrainData$ from $N = 125$ to $N = 4000$.  Because we expect that results will be quite noisy for smaller $\setsize{\TrainData}$, we subsample 20 times for each combination of target task training set size, patient representation, and target task.  We evaluate the effectiveness of the patient level representations using the 34,525 test set patients, with inputs and labels derived from their observation and follow-up eras respectively.  Note that we use stratified sampling to construct the $\TrainData$ with a fixed prevalence of 20\% because the target task positive cases are quite rare.  However, the evaluation is conducted using the natural prevalence of the target task positive cases found in the test set patients. Note that this design does not estimate variance in the performance estimates due to sampling of the test set.  

\subsection{Learning Patient Level Representations of Clinical Text}
In this section, we present two simple methods for representing clinical notes that use the training data to learn efficient patient level representations from clinical notes.  These methods can be framed as \emph{transfer} learning approaches, in which we use a \emph{source task} that uses a large dataset, $\AllData$, to learn how to transform raw patient data into a form that is useful for the source task, i.e., $G: \PatientData \rightarrow \PatientRepr \in \R^d$, and then use $\PatientRepr$ for  target tasks we care about but for which labeled training data is limited. 

\subsubsection{Embed-and-aggregate}
In this approach, we construct $G$ in two phases.  First, as described above, we learn an embedding $e_w \in \R^d$ for each word $w \in \Vocab$.  We then apply arithmetic functions -- mean, max or min -- element-wise to these vectors to aggregate the words in each note to arrive at a note level representation, and  likewise aggregate the note level representations to arrive at a patient level representation: 
\[ \PatientEA =  \underset{j = 1 \dots T^{(i)}}{Aggr} \underset{ w \text{ in note } j}{Aggr} \Embd \]
where $Aggr$ is an element-wise min, max or mean. Note that both phases are unsupervised; that is, no external labels are needed.  In this work, we use the GloVe method for learning these embeddings from the clinical notes from the observation eras of the training set patients.  Embeddings learned by methods such as GloVe and word2vec capture the intuition that semantically similar words occur in similar contexts.  In text, context may simply mean distance in the sequence of words.  However, we are starting from a bag of words representation of each note, and the relative ordering of words within notes is lost.  We thus use an approach used for learning embeddings of medical codes \citep{choi2015doctor, choi2016learning}, which are naturally unordered, and represent each note in the training corpus as 2 random permutations of its words.  We ran GloVe on this corpus with a neighborhood size of 10 for 25 iterations.  

To test whether embeddings learned on different corpora differ in their utility, we also evaluated embeddings learned from 350,000 biomedical research article abstract \citep{de2014medical}.  Terms in the abstracts were mapped to UMLS CUIs, and embeddings were learned using the skipgram word2vec model with a neighborhood size of 5 and dimensionality of 200.  Embeddings for 52,000 CUIs are publicly available,\footnote{\url{https://github.com/clinicalml/embeddings}} and approximately 28,000 of those are in our vocabulary.  We refer to these embeddings as MCEMJ (Medical Concept Embeddings from Medical Journals).  Finally, we must specify the dimensionality of the embeddings; we present experiments using 300 and 500 dimensional embeddings.  

\subsubsection{Recurrent Neural Nets}
In this approach, we use recurrent neural nets whose input is a sequence of bag of words vectors representing the clinical notes written about a given patient in that patient's observation era \citep{choi2015doctor}.  These vectors are input in temporal order to the RNN, and notes written on the same day are merged into a single input vector.  The final hidden state is used as the patient level representation for the following source task: predict the diagnosis codes assigned to the patient during their observation era.  The diagnosis codes are derived from the structured data in each patient's observation era by grouping ICD-9 codes according into Clinical Classification Software (CCS) groups \footnote{https://www.hcup-us.ahrq.gov/toolssoftware/ccs/ccs.jsp}.  CCS maps 12,856 diagnosis codes into 254 broad disease categories such as \emph{lung cancer} or \emph{cardiac dysrhythmias}, and have previously been used as supervision labels for the training of recurrent neural nets \citep{choi2015doctor} because it dramatically reduces the label space.  The source task is thus a multi-task sequence classification problem.  

We used \texttt{tanh} activations, dropout with drop probability 0.2, and a dense layer with binary cross-entropies for the multi-task supervision. Networks were trained with RMSProp \citep{tieleman2012lecture} for 100 epochs using a batch size of 32 and maximum sequence length of 50, with longer sequences truncated to keep the most recent notes.  The models were implemented in Keras 1.2.0 \citep{chollet2015keras} with a Tensorflow 0.12.1 backend \citep{tensorflow} on an NVIDIA GeForce GTX TITAN X GPU. 

\subsubsection{Baseline Representations}
We compared the learned representations against the TF-IDF transformed bag of words (TF-IDF BOW), Latent Semantic Analysis (LSA), and Latent Dirichlet Allocation (LDA) representations presented above.  For TF-IDF BOW, we limited the vocabulary to the 15,000 most frequent words in each training subsample.  Word counts were then summed over each patient's input notes.  The final representation consisted of TF-IDF transformed counts of words for each patient, where the term and inverse document frequencies were computed over the entire training corpus (i.e., all notes from the observation eras of the training set patients).  For LSA, we summed the bag of words vectors for all the notes in each training set patient's observation era, and performed a truncated SVD on the resulting \emph{term-patient} matrix [Ref irlba package].  We truncated the SVD to 600 dimensions.  Finally, for LDA, we fit a topic model on the training corpus using Gensim \citep{gensim}. 300 topics were fit over two passes through the training notes.  The model was then used to estimate topic distributions for validation and test set notes.  The topic distributions for each note in a given patient's observation era were aggregated using mean and max pooling.

\subsection{Target Task Models}
The target task models were L2 regularized logistic regression models using the various patient level representations as fixed length inputs.  These models were tuned via 5-fold cross validation on the training subsamples using the glmnet package \citep{glmnet2010} in R \footnote{http://www.R-project.org}.    

\section{Results}
We now present the results of our experiments, in which we compare the target task performance of our learned representations against each other and our baseline representations.  We first focus on results for broad classes of representations before discussing the overall comparison in order to simplify the latter.  In all experiments, we show three \emph{learning curves} -- estimates of the relationship between generalization performance and target task training set size -- for each of the three target tasks of interest and target task training set sizes of ranging from $N = 125$ to $N = 4000$.  The error bars are the standard error of the mean for each estimate across 20 random subsamples of target task training patients as described above.  

\subsection{Performance of the baseline representations}
\begin{figure}[h]
\centering \includegraphics[width=0.9\textwidth]{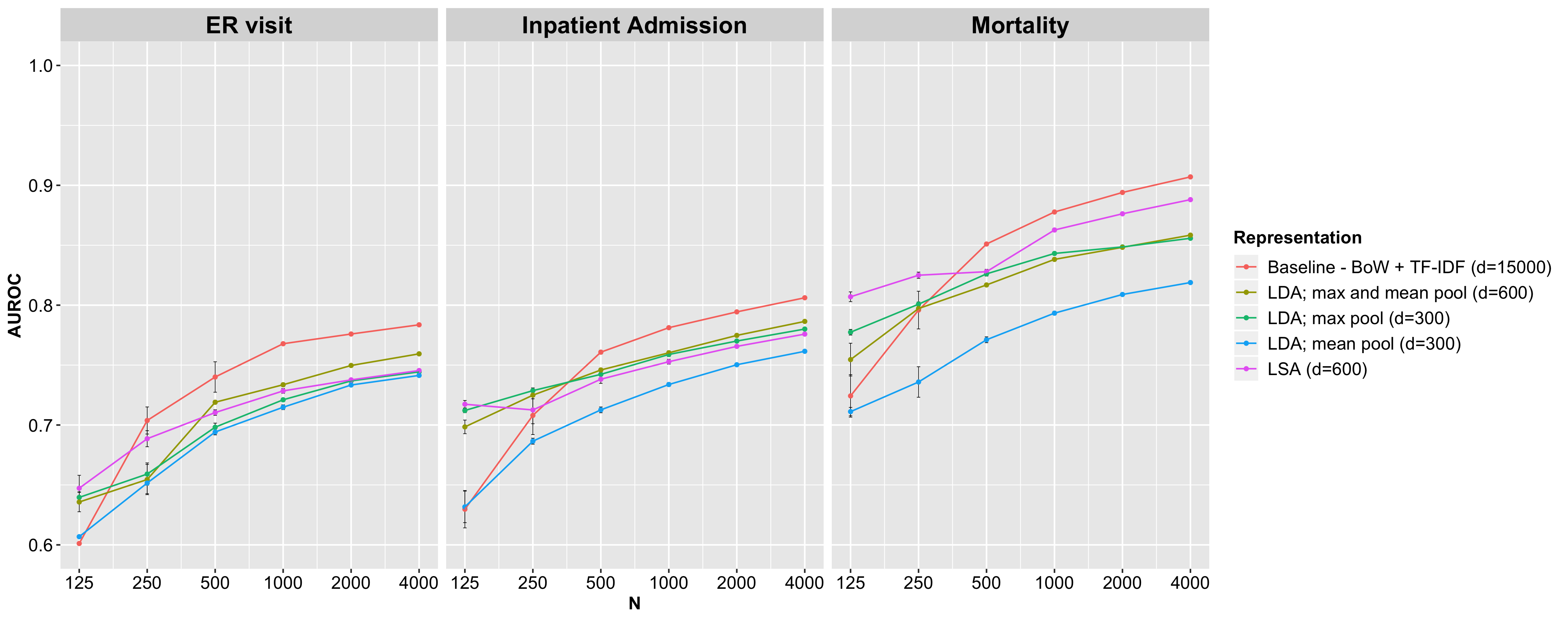}
\caption{Performance of the baseline representations across the three target tasks and a range of training set sizes.       }
\label{fig:results_baselines}
\end{figure}

\autoref{fig:results_baselines} shows the performance of the baseline representations -- bag of words with TF-IDF weighting, LSA, and LDA with mean and max aggregation across notes -- across the target tasks and training set sizes. The bag of words with TF-IDF weighting representation dominates the others at larger sample sizes.  For very small sample size, the LSA representation can improve on TF-IDF weighted bag of words.  Note that mean aggregation for the LDA representation significantly underperforms the other baselines with the exception of the bag of words representation with very small training set sizes, and it appears that max pooling of the topic distributions performs much better.  Generally, it appears that LSA ($\PatientLSA$) does well at very small training set sizes but are quickly overtaken by the simple $\PatientBOW$ representation as training data is added.  We thus focus only on the $\PatientBOW$ and $\PatientLSA$ as baselines representations; it appears that these are both better choices than the simple topic modeling applied to summarize clinical text in Miotto et al \citep{miotto2016deep}.  

\subsection{Embed-and-aggregate}
Embed-and-aggregate method uses simple functions to form a patient level representation from the embeddings of the words found in the patient's notes:  
\[ \PatientEA =  \underset{j = 1 \dots T^{(i)}}{Aggr} \underset{ w \text{ in note } j}{Aggr} \Embd \]
In order to apply this method, we must make the following decisions.  First, what should we use for $Aggr$?  Prior work has suggested that the common choice of mean aggregation may not be optimal.  Second, what should the dimensionality of the embeddings be?  We are interested in this because more aggressive dimensionality reduction often benefits tasks with very small training sets.  And finally, what text should we use to learn the embeddings from?  The most obvious choice is to learn them from the corpus of clinical notes in our dataset, but more useful embeddings might be learned from less noisy text such as the abstracts of biomedical journal articles.  

\subsubsection{Embed-and-aggregate - how should we aggregate?}
\begin{figure}[h]
\centering \includegraphics[width=0.9\textwidth]{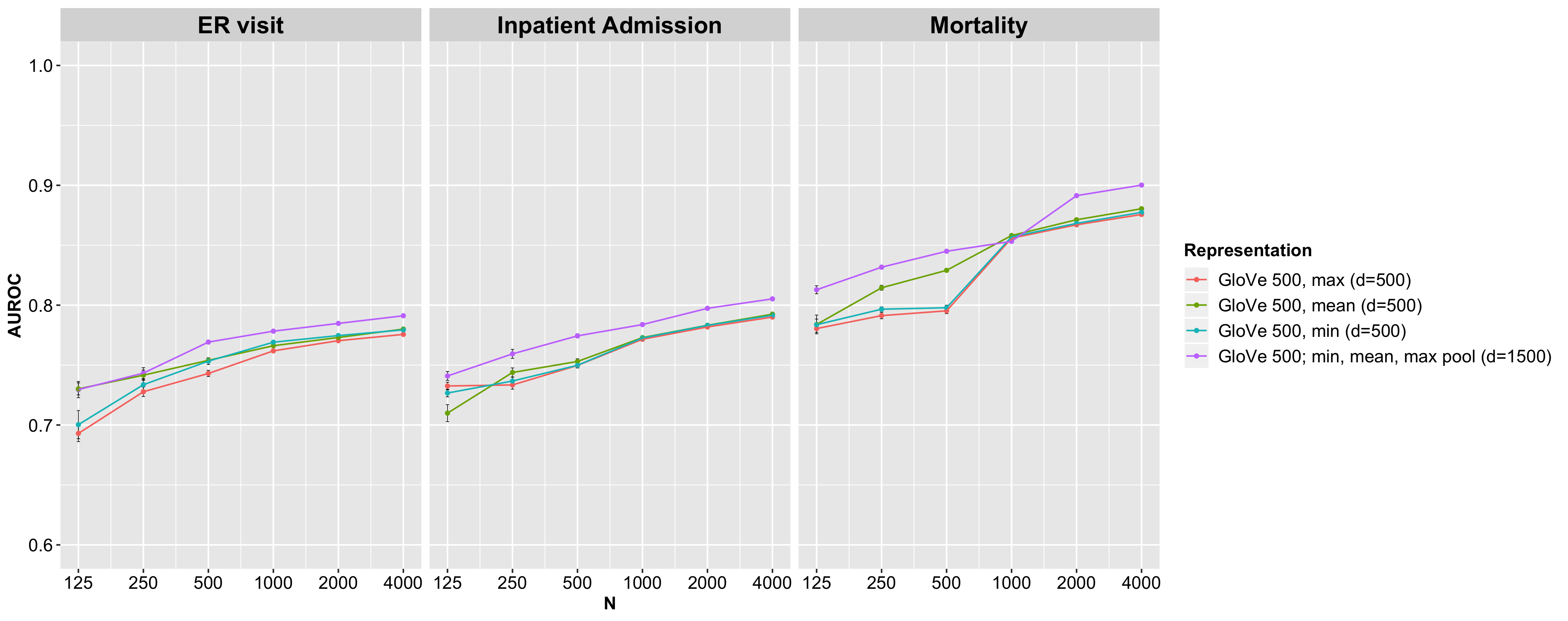}
\caption{Performance of Embed-and-aggregate using $d=500$ embeddings learned from clinical notes and element-wise \emph{min}, \emph{max}, and \emph{mean} as aggregation functions.  We also test concatenation of the resulting patient level representations for all of these choices.}
\label{fig:EA_which_aggr}
\end{figure}

 \autoref{fig:EA_which_aggr} shows results for aggregating embeddings using $min$, $max$, and $mean$, along with concatenating the patient level embeddings for each of these choices into a single, larger patient level representation.  The patient level representation resulting from concatenating the \emph{min}, \emph{max} and \emph{mean} aggregated patient level representations dominate all other choices for all targets and over all target task training set sizes.  Interestingly, this holds even for the smallest target task training set size of $N = 125$ despite the much higher dimensionality of the concatenated representation ($d = 1500$ for the latter vs $d = 500$ for the other choices).  

\subsubsection{Embed-and-aggregate - Dimensionality}
\begin{figure}[h]
\centering \includegraphics[width=0.9\textwidth]{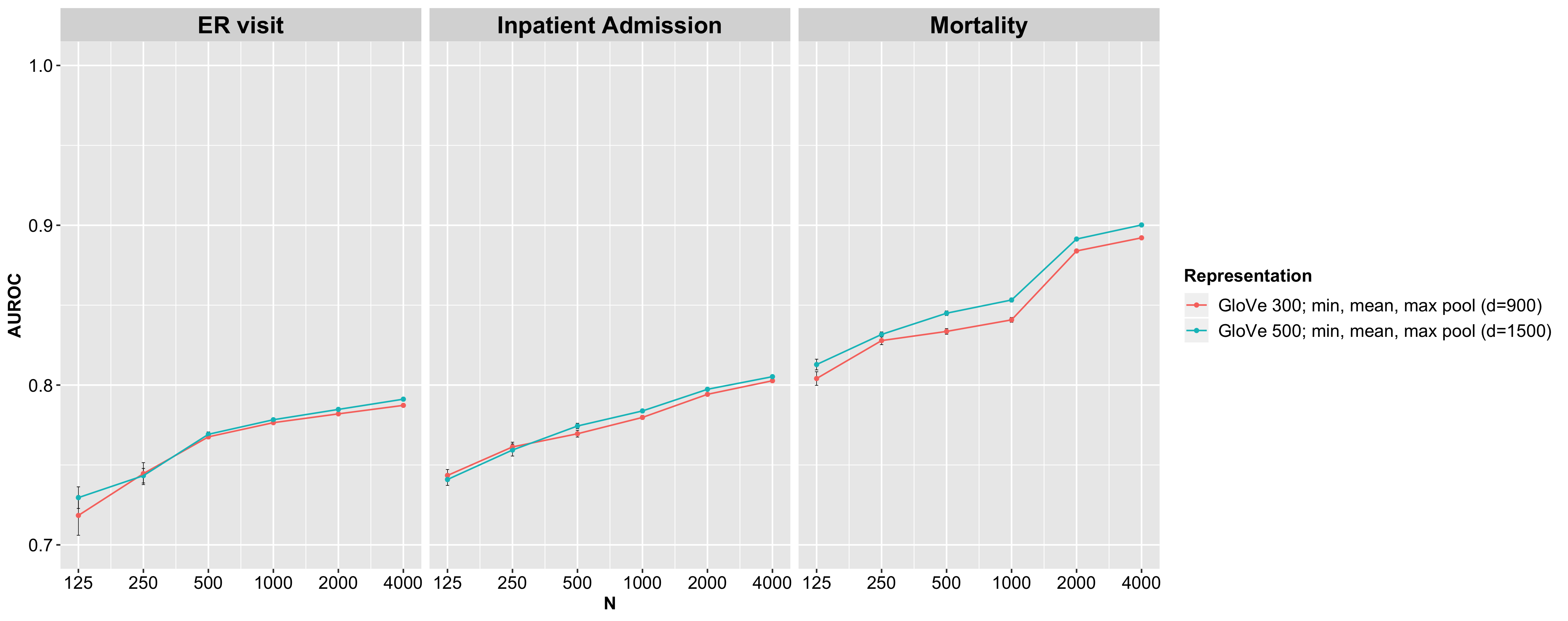}
\caption{Performance of Embed-and-aggregate using $d=300$ vs $d=500$ embeddings learned from the text of notes from the observation eras of training set patients. }
\label{fig:EA_dims}
\end{figure}

In this experiment, we assess the impact of the dimensionality of the embeddings learned from the clinical text on target task performance.  \autoref{fig:EA_dims} suggests that there is little penalty incurred by using the higher dimensional embeddings for the three target tasks, even for very small target task training set sizes ($N = 125)$.  In fact, for the Mortality target task, the higher dimensional representation dominates the lower dimensional representation across the range of target task training set sizes.  

\subsubsection{Embed-and-aggregate - which text should we learn from?}
\begin{figure}[h]
\centering \includegraphics[width=0.9\textwidth]{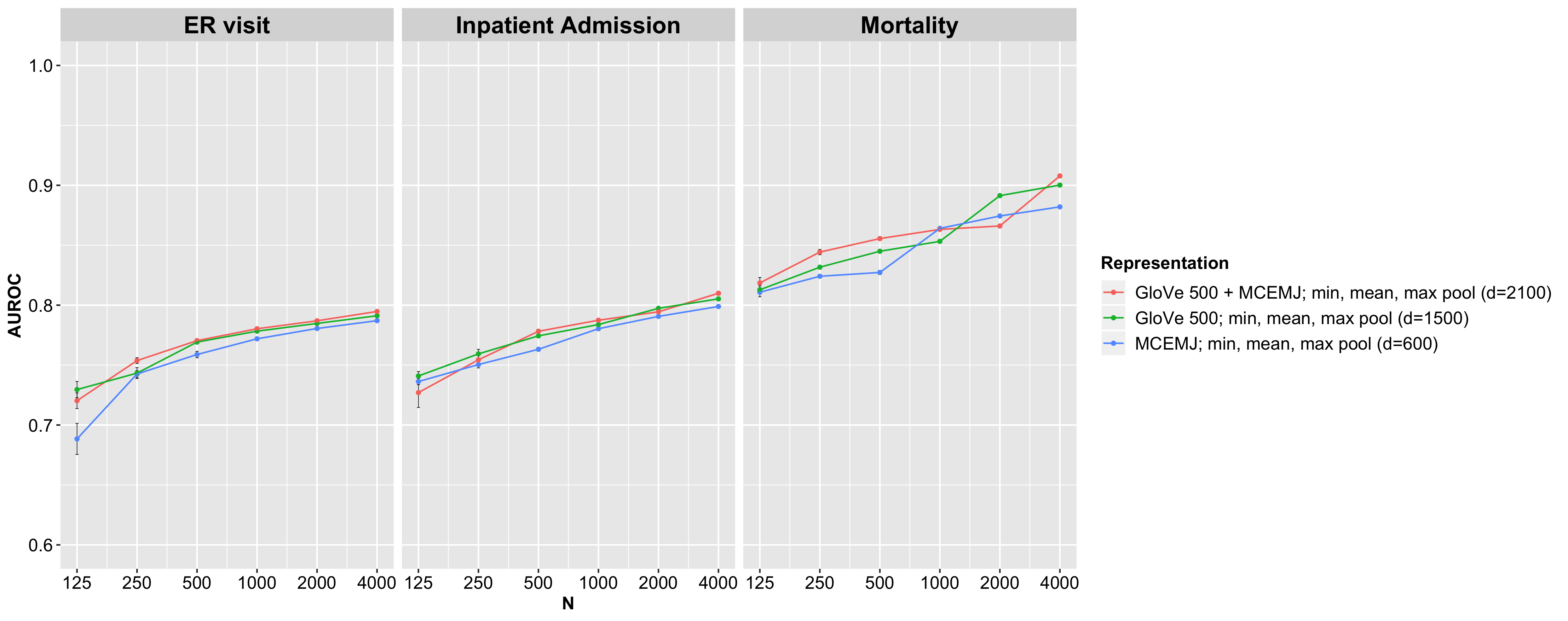}
\caption{Performance of Embed-and-aggregate using embeddings learned from the text of clinical notes using GloVe vs the abstracts of biomedical research articles (MCEMJ). }  
\label{fig:EA_which_text}
\end{figure}

While we have a large corpus of clinical notes from which to learn embeddings, it is not obvious that this is the best corpus from which to learn embeddings.  Clinical notes are noisy, and our text processing pipeline, which reduces each note into a bag of words, may be losing valuable information.  We therefore assess the relative utility of embeddings learned from an alternative source of text -- the abstracts of 350,000 biomedical journal articles.  In \citep{de2014medical}, the word2vec skipgram model is used to learn $d=200$ dimensional embeddings for 58,000 biomedical concepts (CUIs).  Of these, 28,000 are in our vocabulary.  \autoref{fig:EA_which_text} shows the performance of Embed-and-aggregate on the target tasks using embeddings learned from the clinical text vs journal article abstracts ("MCEMJ").  In both cases, we use the the concatenation of \emph{min}, \emph{max}, and \emph{mean} aggregated embeddings.  

We also consider embeddings that are the concatenation of those learned from the clinical text and research article abstracts.  Here, there is no clear winner: for the most part, the MCEMJ embeddings perform somewhat worse than the other choices, but this may reflect the lower dimensionality (recall that higher dimensional embeddings learned from clinical text perform better than lower dimensional embeddings learned from the same corpus even for very small target task training sets).  Note that the combined embeddings generally fare no worse than the other options, particularly for smaller target task training sets, suggesting that the two corpora provide somewhat complementary information.  However, differences in the methods used to learn the embeddings, along with the different sized corpora and dimensionality of the embeddings render these results difficult to interpret.  We therefore focus on the concatenation of \emph{min}, \emph{max}, and \emph{mean} aggregated $d=500$ dimensional embeddings learned from clinical text, and use \emph{Embed-and-aggregate} or $\PatientEA$ to refer to this specific configuration.  

\subsection{RNNs - preliminary tuning}
We performed preliminary experiments using the validation split in order to assess various technical aspects of the RNN models.   Below we briefly discuss the main conclusions.  

\begin{itemize}
\item We found that providing supervision to the RNN after the final timestep was more effective than providing supervision at each time step.  

\item Typically, the first layer of RNNs with bag of words inputs multiply an embedding matrix by the input vector, effectively summing the embeddings for the inputs.  However, we found that applying an element-wise maximum to the embeddings at each time-step significantly improved performance in the target tasks.    

\item We explored initializing the embedding layer of our recurrent nets using word embeddings learned from the clinical notes or from from research article abstracts.  We found that random initialization consistently outperformed initialization with pre-trained embeddings. 

\item Finally, we explored using Long Short-Term Memory (LSTM) \citep{hochreiter1997long} and Gated Recurrent Units (GRU) \citep{cho2014learning} as recurrent cells.  We found that, consistent with \citep{choi2015doctor}, GRUs consistently outperformed LSTMs.  Thus, our final experiments used GRUs with max aggregation of word embeddings.
\end{itemize}

\subsection{RNNs - Dimensionality}
\begin{figure}[h]
\centering \includegraphics[width=0.9\textwidth]{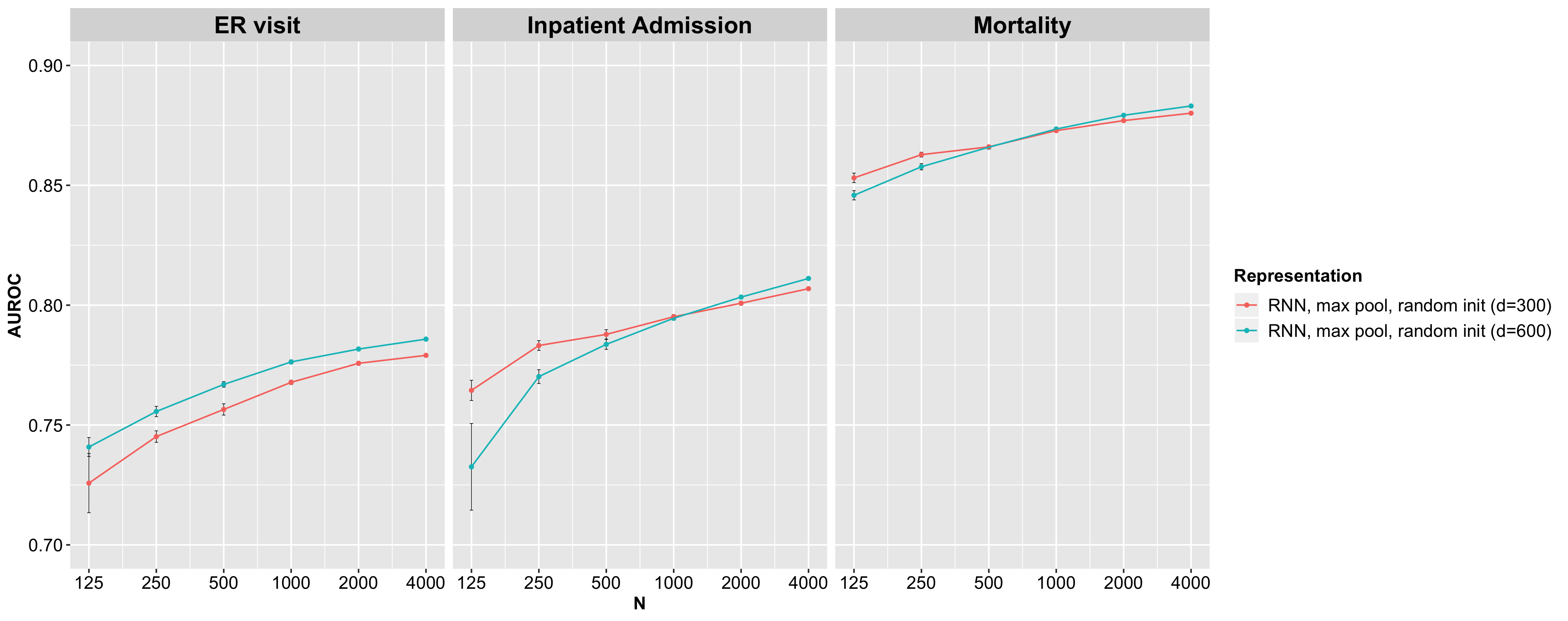}
\caption{RNN derived representations with $d = 300$ vs $d = 600$.}
\label{fig:RNN_dims}
\end{figure}

Recurrent neural nets have a vast array of hyperparameters that can have a significant impact on their performance in various tasks.  However, tuning RNNs is computationally expensive, and at any rate we are interested in using RNNs as fixed feature extractors after fitting them on the source task.  It is not clear that incrementally better performance on the source task will lead to better performance on target tasks with severely limited training data.  Reflecting these realities, we did not conduct an extensive hyperparameter search to optimize the performance of our RNNs on the source task; rather, we focused on broad choices such as where LSTMs or GRUs work better as the recurrent units, and whether mean vs max aggregation works better, as described above.  We then turned to the dimensionality of the learned representation, evaluating models with $d = 300$ and $d = 600$.  \autoref{fig:RNN_dims} shows the results of that evaluation.  For one of the targets, future ER visits, the higher dimensional representation works better even for $N = 125$.  For the other two targets, however, we observe the expected pattern -- the lower dimensional ($d=300$) representation is somewhat better with very small target task training sets, but as the target task training set grows the higher dimensional representation catches up and does well. Because it is not clear which representation is best, we use both in our overall evaluation below.  

\subsection{Overall}
\begin{figure}[h]
\centering \includegraphics[width=0.9\textwidth]{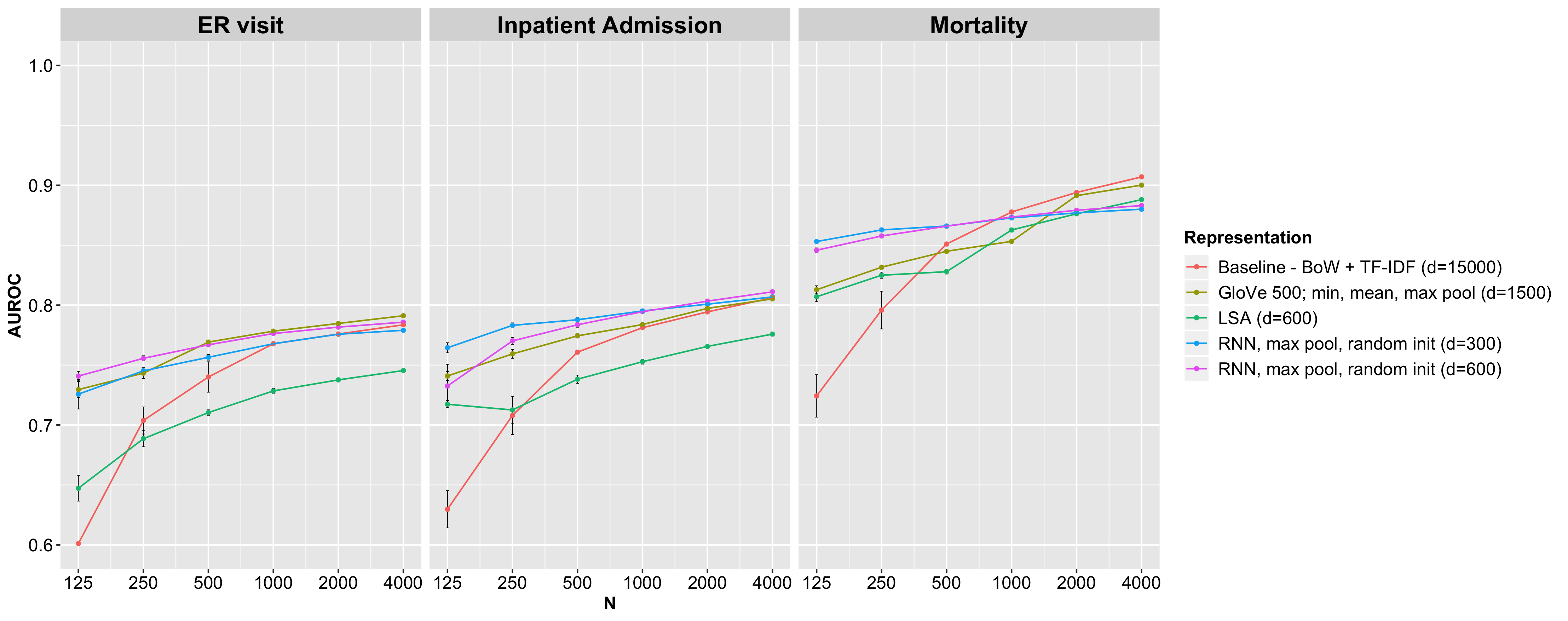}
\caption{Overall comparison of the best baseline models vs the best learned representations. } 
\label{fig:overall_results}
\end{figure}

Of course we must ask how well our learned representations fare against the baselines of bag of words with TF-IDF weighting and LSA.  \autoref{fig:overall_results} shows the results for the best overall \emph{embed-and-aggregate} representation, along with our two RNN based representations.  As expected, the learned representations are significantly better with smaller target task training sets, outperforming the baselines by substantial margins.  The RNN representations generally beat the \emph{embed-and-aggregate} representation, though the latter comes quite close to the two RNN representations, especially as the target task training set grows.  However, the most notable pattern is that with sufficiently large target task training sets, the humble bag of words representation catches up to, and in the case of Mortality eventually surpasses, the performance of the best learned representations.  

\subsection{Wide and deep representations}
\begin{figure}[h]
\centering \includegraphics[width=0.9\textwidth]{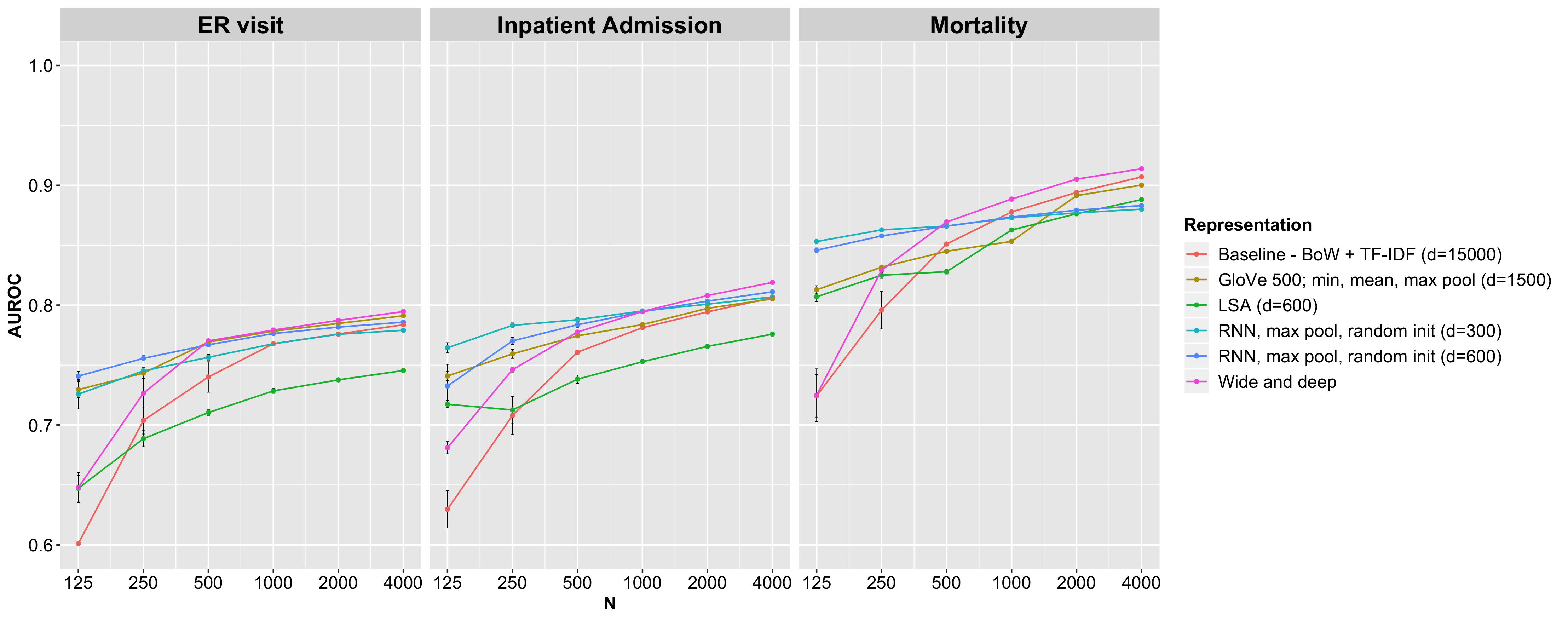}
\caption{ Performance of a \emph{wide and deep} representation. }
\label{fig:wide_and_deep}
\end{figure}

The strong performance of the learned representations with small $\TrainData$, along with the strong performance of the simple bag of words representation with large $\TrainData$, begs the question of whether it is possible to combine their benefits.  One simple approach for attempting this is to concatenate the two representations so that the models for the target tasks have access to both the reduced dimension summary provided by the learned representations and the fine grained information of the bag of words representation.  This is roughly analogous to the so-called \emph{wide and deep} representations that have been used for problems with very high dimensional and sparse inputs \citep{widedeep}.  \autoref{fig:wide_and_deep} shows the results when we combine the $d=600$ RNN representation with the bag of words.  Interestingly, adding the summary information provided by the learned representations to the bag of words representation strictly improves performance across the range of target task training set sizes, with the wide and deep representations out-performing the bag of words representations by significant margins at most training set sizes (the one exception is $N = 125$ for the Mortality task). The gain in performance with small target task training sets is not great enough to catch up to or surpass the learned representations; the extra information is still difficult to exploit due to the curse of dimensionality.  Nevertheless, these results suggest that this approach may yield significant improvements over BOW and the learned representations alike when training data is plentiful.  Of course, with enough training data we might be best off learning models for the target task directly instead of resorting to transfer learning.  

\section{Discussion}
We have demonstrated that learned representations of clinical text can be effective for predicting clinical events in scenarios where labeled data for supervised learning is scarce.  Our learned representations range from a very simple \emph{embed-and-aggregate} method to a more computationally intensive method that uses recurrent neural nets.  We find that embed-and-aggregate is nearly as effective as the RNN based method and is much simpler to compute and share.  In comparison to various commonly used baselines, we find that our learned representations admit significant more accurate predictive models with very small target task training set sizes.  However, as the training set size available for supervised learning increases, we find that the simple bag of words with TF-IDF weighting can close the performance gap, or sometimes even surpass the performance of the learned representations.  In such settings, it may be possible to do even better than bag of words and the learned representations by simply concatenating the two representations together.  Of course, with sufficient training data it may be possible to learn a complex model for the target task directly.  

We acknowledge several important limitations of this study.  First, we note that we have started all of our methods from the same bag of words representation of the clinical notes, since that is the output of our text processing pipeline.  It is of course possible that there is much to be gained by learning representations directly from the full text of the notes.  Second, we note that the way we have chosen the prediction time for each patient does not correspond to the way that predictions would be made at run time since we have picked the prediction time in reference to an event that in the future relative to the prediction time.  This likely induces a somewhat optimistic bias in the performance estimates.  However, we believe that our conclusions regarding the \emph{comparative efffectiveness} of the different patient level representations are still valid, though of course follow up work with more careful problem formulation is warranted.  Third, we must note that we have constructed our target tasks to be experimentally convenient but at the cost of realism in important respects.  For each of the target tasks, we actually have ample training data since the positive cases are not too rare and are easily ascertained from EHR data.  In addition, in a real world setting with a truly limited labeled dataset for the target task of interest, our estimates of generalization error would be much noisier because the evaluation sets would likely be much smaller than that used in our study.  This implies that we would be hard pressed to accurately select the representation that truly admits the best generalization performance.  These limitations notwithstanding, our results suggest that relatively simple methods drawn from NLP can yield substantial performance gains over commonly used baseline representations of clinical text.  

\section*{References}
\bibliographystyle{plainnat}
\bibliography{ref}

\end{document}